\pdfoutput=1

\documentclass[11pt]{article}

\usepackage{EMNLP2023}

\usepackage{times}
\usepackage{latexsym}

\usepackage[T1]{fontenc}

\usepackage[utf8]{inputenc}

\usepackage{microtype}

\usepackage{inconsolata}
\usepackage{graphicx}
\usepackage{inconsolata}
\usepackage{array}
\usepackage{mathrsfs}
\usepackage{amsthm}
\usepackage{amsmath}
\usepackage{amsfonts}

\usepackage[noend]{algpseudocode}
\usepackage{multirow}
\usepackage{booktabs}
\usepackage{color}
\usepackage{arydshln}
\usepackage{makecell}
\usepackage{soul}
\usepackage{cleveref}
\crefname{section}{§}{§§}
\Crefname{section}{§}{§§}

\usepackage{lipsum}
\newcommand\blfootnote[1]{
  \begingroup
  \renewcommand\thefootnote{}\footnote{#1}
  \addtocounter{footnote}{-1}
  \endgroup
}

\title{E\textsc{xplain}, E\textsc{dit}, G\textsc{enerate}: Rationale-Sensitive Counterfactual Data Augmentation for Multi-hop Fact Verification}

\author{Yingjie Zhu\textsuperscript{\rm 1*}, Jiasheng Si\textsuperscript{\rm 1*}, Yibo Zhao\textsuperscript{\rm 1}, Haiyang Zhu\textsuperscript{\rm 1}, Deyu Zhou\textsuperscript{\rm 1$\dagger$}, Yulan He\textsuperscript{\rm 2,3} \\
     \textsuperscript{\rm 1}School of Computer Science and Engineering, Key Laboratory of Computer Network \\ and Information Integration, Ministry of Education, Southeast University, China\\
     \textsuperscript{\rm 2} Department of Informatics, King’s College London, UK\\
     \textsuperscript{\rm 3} The Alan Turing Institute, UK\\
    \texttt{\{yj\_zhu, jasenchn, yibozhao, haiyangzhu, d.zhou\}@seu.edu.cn} \\
    \texttt{yulan.he@kcl.ac.uk} \\
    }

\begin{document}
\maketitle
\blfootnote{\textsuperscript{*}Equal Contribution.}
\blfootnote{\textsuperscript{$\dagger$}Corresponding Author.}
\begin{abstract}
Automatic multi-hop fact verification task has gained significant attention in recent years. Despite impressive results, these well-designed models perform poorly on out-of-domain data. One possible solution is to augment the training data with counterfactuals, which are generated by minimally altering the causal features of the original data. However, current counterfactual data augmentation techniques fail to handle multi-hop fact verification due to their incapability to preserve the complex logical relationships within multiple correlated texts.
In this paper, we overcome this limitation by developing a rationale-sensitive method to generate \textit{linguistically diverse} and \textit{label-flipping} counterfactuals while preserving \textit{logical relationships}. In specific, the diverse and fluent counterfactuals are generated via an Explain-Edit-Generate architecture. Moreover, the checking and filtering modules are proposed to regularize the counterfactual data with logical relations and flipped labels.  Experimental results show that the proposed approach outperforms the SOTA baselines and can generate linguistically diverse counterfactual data without disrupting their logical relationships\footnote{The code and datasets are available at \url{https://github.com/AAAndy-Zhu/RACE}}.

\end{abstract}

\section{Introduction}
\label{sec:introduction}

Multi-hop fact verification task, which discerns the truth from falsehood based on multiple hops of reliable evidence, becomes crucial in countering misinformation and counterfeit news spread on current social media platforms \citep{vosoughi-2018, brenda-botnevik-2020},
especially in some specific domains such as politics \citep{ liar-plus, politihop}, science \citep{scifact, wadden-etal-2022-multivers} and public health \citep{pubhealth, healthver}. 
However, many recent works often perform poorly under the multitude of distribution shifts due to an over-reliance on spurious correlations between input text and labels~\citep{gururangan-etal-2018-annotation, schuster-etal-2019-towards, geirhos2020shortcut}. It can potentially be addressed by Counterfactual Data Augmentation (CDA), using counterfactual instances generated by perturbing causal features within the input~\citep{khashabi-etal-2020-bang}. Several works have revealed that training with counterfactual data enhances the capability of the model to identify causal features and diminish its reliance on spurious correlations between the input text and the label, thus resulting in the improvement in Out-Of-Domain (OOD) generalization \citep{vig2020causal, eisenstein-2022-informativeness}.

\begin{table*}[t]
    \centering
    \resizebox{16cm}{!}{
    \renewcommand\arraystretch{1.1}
\begin{tabular}{ccp{15cm}}
\toprule[2pt]

 \textbf{Task} & \textbf{Inference} &  \multicolumn{1}{c}{\textbf{Input X (label Y)}} \\\midrule[1pt]

  \begin{tabular}[c]{@{}c@{}}Sentiment Analysis\end{tabular} 
&  \begin{minipage}[c]{0.21\columnwidth}
		\centering
		\includegraphics[width=\linewidth]{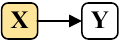}\end{minipage}         
&  This is an \textcolor[RGB]{192,0,0}{amazing} book, I'm already immersed in the storyline. \textbf{(POSITIVE)}        \\\midrule[0.5pt]

 \multirow{2}{*}{\begin{tabular}[c]{@{}c@{}}Single-hop\\ Fact Verification\end{tabular}} 
& \multirow{2}{*}{
\begin{minipage}[c]{0.38\columnwidth}
		\centering
		\includegraphics[width=\linewidth]{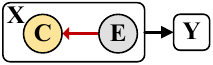}\end{minipage} 
}     
&\textbf{C:} Little Miss Sunshine was filmed \textcolor[RGB]{192,0,0}{over 30 days}. \textbf{(SUPPORTS)} \\
\cdashline{3-3}

 &  & \textbf{E:} Little Miss Sunshine ..., filming began on June and took place \textcolor[RGB]{192,0,0}{over 30 days} in Arizona ...  \\\midrule[0.5pt]

 \multirow{5}{*}{\begin{tabular}[c]{@{}c@{}}Multi-hop\\ Fact Verification\end{tabular}} & \multirow{5}{*}{
\begin{minipage}[c]{0.4\columnwidth}
		\centering
		\includegraphics[width=\linewidth]{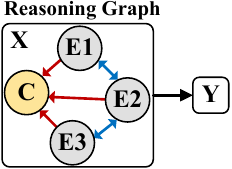}\end{minipage} 
}     
&\textbf{C:} The Ford Fusion was introduced for model year 2006. The Rookie of The Year in the 1997 CART season drives it in the NASCAR Sprint Cup Series. \textbf{(SUPPORTS)}\\
\cdashline{3-3}

  &   & \textbf{E1:} \textcolor{blue}{Ford Fusion} is manufactured and marketed by Ford. \textcolor[RGB]{192,0,0}{Introduced for the 2006 model year}, …      \\

      &               & \textbf{E2:} \textcolor{blue}{Patrick Carpentier} competed \textcolor[RGB]{192,0,0}{in the NASCAR Sprint Cup Series}, \textcolor[RGB]{192,0,0}{driving the} \textcolor{blue}{Ford Fusion}.                         \\

 &           &   \textbf{E3:} \textcolor[RGB]{192,0,0}{The 1997 CART} PPG World Series season, ... \textcolor[RGB]{192,0,0}{Rookie of the Year} was \textcolor{blue}{Patrick Carpentier}.          \\
\bottomrule[2pt]
\end{tabular}}
\caption{Comparison between different tasks.}
\label{tab:introduction}
\end{table*}

In this paper, we seek to generate counterfactuals for multi-hop fact verification, instead of exploring the causal bias for a specific model.
However, due to the complex logical relationships within the multi-hop input texts, developing such an approach poses some significant challenges.
As shown in the first row of Table~\ref{tab:introduction}, most CDA methods are designed for NLP tasks without requiring intricate reasoning over the input, such as the sentiment analysis task \citep{yang-etal-2021-exploring, howard-etal-2022-neurocounterfactuals}. 
Their local modification of the causal feature in a single sentence (e.g., ``\textit{amazing}'' in Table \ref{tab:introduction} $\Rightarrow$ ``\textit{terrible}'') is difficult to constrain the \textit{logical relationships} between different causal features in multiple correlated texts,
resulting in unverifiable counterfactuals.
Furthermore, the prior attempt, CrossAug \cite{lee2021crossaug}, is primarily designed to generate counterfactuals for single-hop fact verification via consistently editing the causal features in the claim and in the one piece of evidence (e.g., ``\textit{over 30 days}
'' in the second row of Table \ref{tab:introduction} $\Rightarrow$ ``\textit{less than 10 days}''). Nevertheless,
its claim-only based generation strategy struggles to preserve the complex logical relationships when faced with multiple hops of evidence, and fails to ensure \textit{label flipping} and \textit{linguistic diversity} in the counterfactuals,
which are crucial for CDA~\cite{joshi-he-2022-investigation}.

For multi-hop fact verification, as shown in the third row of Table \ref{tab:introduction},
the set of possible causal features is more complex, and exploring them may necessitate intricate reasoning about the logical relationships between multiple hops of evidence and between the claim and the evidence.
For example,
the ``\textit{Patrick Carpentier}'' in $E2$, which is invisible to the claim, bridges the connection between the causal features ``\textit{Introduced for the 2006 model year}'' in $E1$ and  ``\textit{Rookie of the Year}'' in $E3$, thus leading to the alignment of the multi-hop evidence with the claim $C$ (as shown in the Reasoning Graph).
Without considering such complex \textit{logical relationships} within the correlated input, 
the generated counterfactual claims potentially tend to be unreasonable or unverified.
Furthermore, 
ensuring the \textit{label flipping} and \textit{linguistic diversity} of generated counterfactuals become increasingly difficult with the premise of \textit{logical relationships}, which are critical factors to assure the quality of the counterfactuals.

To address these challenges, we develop a novel pipeline method, RACE (\textbf{RA}tionale-sensitive \textbf{C}ounterfactual g\textbf{E}neration),
by focusing on the causal features within the rationales extracted from the multi-hop evidence using an explainability method. In specific, for each original instance, the \textit{Explainer} and \textit{Editor} modules are employed to produce the counterfactual evidence that logically corresponds to --- but factually distinct from --- the original claim. Then, according to the counterfactual evidence, an entity-aware \textit{Generator} generates the counterfactual claims by synthesizing the semantic information across multi-hop evidence.
During the above process, the Checking and Filtering modules are used to regularize the reasonableness of the output of each module from different aspects, resulting in fully labeled examples that can be used directly to augment the training data. The \textbf{motivation} here is that these rationales provide the intrinsic semantic and relational information for inferring its label, and present the factual consistency with its claim \cite{raha2023neural}.

It should be pointed out that RACE requires no external knowledge as used in \citet{paranjape-etal-2022-retrieval} besides the original training data, and is able to generate \textit{linguistically diverse} and \textit{label-flipping} counterfactuals while preserving \textit{logical relationships}. Compared to alternative approaches (e.g., ChatGPT \cite{chatgpt}) (\cref{sec:experiments}), training on the counterfactuals generated by RACE reveals the improvement in performance under different settings (\cref{subsec:main_results}),
including in-domain, out-of-domain \cite{paranjape-etal-2022-retrieval}, and challenge settings \cite{gardner-etal-2020-evaluating}. 
In addition, the intrinsic evaluation shows that the counterfactual claims generated by RACE are more logical and linguistically diverse than those produced by the baselines (\cref{subsec:intrinsic_evaluation}, \cref{subsec:qualitative_evaluation}). Finally, we compare the results based on different generation models with baselines, illustrating that our method is generation model-agnostic (\cref{subsec:generation_models}).

\begin{figure*}[t]
    \centering
    \includegraphics[width=15cm]{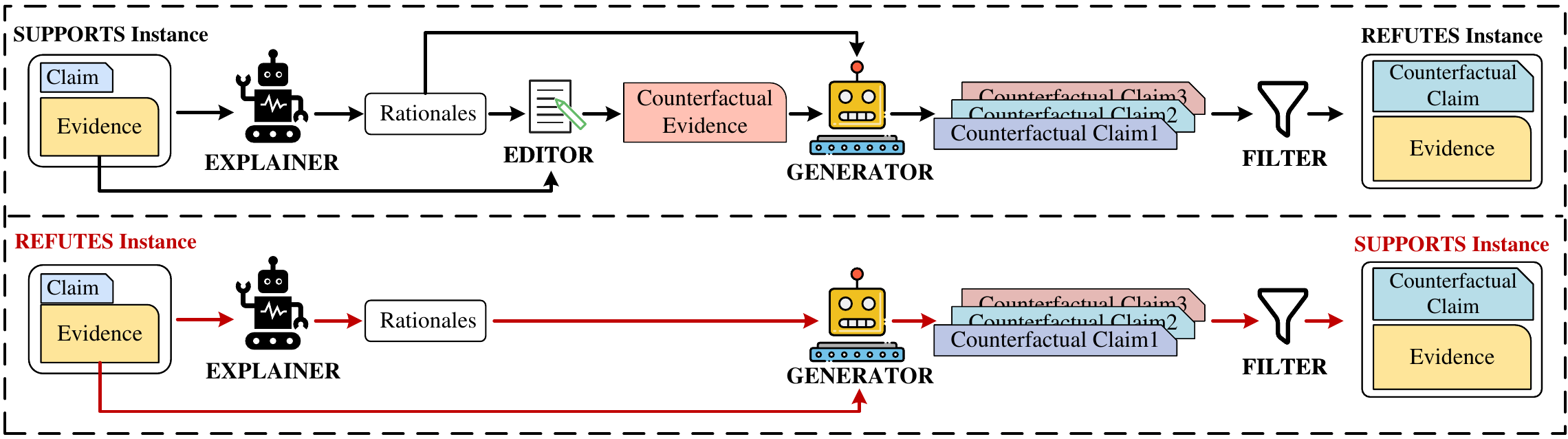}
    \caption{The overall pipeline of RACE. The \textit{SUPPORTS} and \textit{REFUTES} instances are processed differently, as indicated by the black and \textcolor[RGB]{192,0,0}{red} arrows, respectively.}
    \label{fig:method}
\end{figure*}

\section{Related Works}
\paragraph{Debiasing Fact Verification}

A variety of advanced multi-hop fact verification methods have recently emerged in various domains due to the development of pre-trained models~\citep{das2023state}. 
Nevertheless, most models exhibit poor OOD generalization, primarily due to their over-reliance on spurious correlations between inputs and labels \citep{gururangan-etal-2018-annotation, schuster-etal-2019-towards, geirhos2020shortcut}.
Thus, several works focus on the debiasing of fact verification models. 
\citet{schuster-etal-2019-towards} have identified strong cues for predicting labels solely based on the claim.
\citet{zhu2022generalizing} proposed an entity debiasing framework that mitigates entity bias from a cause-effect perspective. 
\citet{lee2021crossaug} addressed the debiasing of fact verification models by augmenting the data with contrastive instances. 
\citet{atanasova-etal-2022-fact} explored what information is sufficient to verify a claim, and proposed a CDA schema for learning of (in)sufficient information.

\paragraph{Counterfactual Data Augmentation}
There is a growing academic interest in CDA to improve model robustness.
Initial studies focus on human-crafted counterfactuals \citep{Kaushik2020Learning, gardner-etal-2020-evaluating}. 
Recently, numerous automatic CDA methods have been proposed for sentiment analysis \citep{wang2021robustness, yang-etal-2021-exploring, howard-etal-2022-neurocounterfactuals}, question answering \citep{paranjape-etal-2022-retrieval, dixit-etal-2022-core}, and natural language inference \citep{glockner-etal-2018-breaking}. However, these methods are primarily targeted to NLP tasks without requiring complex reasoning about the input. Thus, their direct application to the multi-hop fact verification task presents considerable challenges.

\section{Methodology}

Given a claim $c$ with its associated evidence $E=(e_1,e_2,\dots,e_n)$, the aim of multi-hop fact verification is to infer whether the claim is \textbf{supported} or \textbf{refuted} by the evidence. 
We denote an instance in the dataset $D$ as a triplet $(c, E, y)$, where $y\in\{SUP, REF\}$ is the verification label.
The goal of RACE is to generate counterfactual data $(c', E, y')$ or $(c, E', y')$ that differ in some meaningful way from the original instance $(c, E, y)$, where $y' \neq y$,
$c'$ and $E'$ denote the counterfactual claim and counterfactual evidence, respectively. 
This setting poses some unique challenges,
such as requiring to identify the causal features to be edited,
ensuring sound logical relations in evidence editing and claim generation,
and avoiding unverifiable claims.
Meanwhile,
ensuring the semantic diversity and the minimal perturbation of the counterfactuals can also be challenging.
To this end, we propose a general pipeline, RACE, to tackle these challenges.

As shown in Figure \ref{fig:method}, our RACE consists of four stages: (\uppercase\expandafter{\romannumeral1}) Explainer: rationale extraction (\S\ref{sec:rationale extraction}), (\uppercase\expandafter{\romannumeral2}) Editor: evidence editing (\S\ref{sec:evidence editing}), (\uppercase\expandafter{\romannumeral3}) Generator: claim generation (\S\ref{sec:claim generation}), (\uppercase\expandafter{\romannumeral4}) Filtering (\S\ref{sec:filtering}). 
Note that our method handles $SUP$ and $REF$ instances differently, 
as the large difference in generation space between these two types of instances. 

\subsection{Explainer: Rationale Extraction}
\label{sec:rationale extraction}
Our RACE focuses on identifying the causal features within rationales that can be perturbed.
To this end, we use CURE \citep{si2023consistent}, a multi-granular rationale extraction method, to simultaneously extract sentence rationales $R_s$ and token rationales $R_t$ from the multi-hop evidence $E$ for both $SUP$ and $REF$ instances. 
In essence, the token rationales $R_t$ reflect the logical correlation within the evidence (\textcolor{blue} {blue words} in Table \ref{tab:introduction}) and the factual relationship between the claim and the evidence (\textcolor[RGB]{192,0,0} {red words} in Table \ref{tab:introduction}).
Considering the causal relationship of the rationales to the prediction label \citep{wu2022discovering}, 
we regard the extracted rationales as the \textbf{causal features} that are to be further processed. 
The detailed algorithm can be found in  \citet{si2023consistent}.

\subsection{Editor: Evidence Editing}
\label{sec:evidence editing}
In general,
entities contained within the multi-hop evidence possess a rich trove of factual knowledge and crucial information (e.g., \textit{date}, \textit{location}, \textit{organization}, \textit{person}, and the correlation between them), facilitating more precise multi-hop fact verification \citep{mentionmemory, rani2023factify}.
Therefore, 
we meticulously design a set of simple entity-based evidence editing rules to control the semantic perturbation while preserving the multi-hop correlation within the evidence,
and an Ad-Checking module to filter out the under-edited or over-edited evidence.
Additionally,
\citet{tan-etal-2023-multi2claim} highlight that controlling the generation for $REF$ is more challenging due to its significantly broader generation scope compared to $SUP$.
As such, we focus on editing the evidence $E$ for instances $(c, E, SUP)$ rather than for instances $(c, E, REF)$.

\paragraph{Editing}
We first utilize an off-the-shelf NER tool, Stanza \citep{qi-etal-2020-stanza}, to identify various types of \textbf{causal entity} $T$ from token rationales $R_t$. 
Following \citet{rani2023factify}, we only retain entities with specific types, including {\fontfamily{qcr}\selectfont ORG}, {\fontfamily{qcr}\selectfont PERSON}, {\fontfamily{qcr}\selectfont DATE}, {\fontfamily{qcr}\selectfont GPE}, and {\fontfamily{qcr}\selectfont NUM}. 
Then, we automatically edit the evidence according to the following rules.

$\bullet$ \textbf{in-Dataset}: Randomly \textit{replace} entities of type {\fontfamily{qcr}\selectfont GPE}, {\fontfamily{qcr}\selectfont DATE} and {\fontfamily{qcr}\selectfont NUM} with other entities of the same type present in the entire dataset, e.g., \textit{2006 model year} $\Rightarrow$ \textit{2008 model year} in Table \ref{tab:introduction}.

$\bullet$ \textbf{in-Instance}: If all the token rationales in evidence $E$ contain two or more {\fontfamily{qcr}\selectfont PERSON}/{\fontfamily{qcr}\selectfont ORG} entities, their positions are randomly \textit{swapped} between different pieces of evidence, e.g., \textit{Ford} ({\fontfamily{qcr}\selectfont PERSON}) $\Leftrightarrow $ \textit{Patrick Carpentier} ({\fontfamily{qcr}\selectfont PERSON}) in Table \ref{tab:introduction}.
 
$\bullet$ \textbf{Consistent Edit}: 
The same entity token is processed consistently throughout all pieces of evidence, to preserve the multi-hop correlation within the evidence.
For example, if an entity is identified in one piece of evidence, it will be consistently replaced or swapped across all pieces of evidence within the instance.

We use the editing rules to produce one edited evidence for each instance based on a random seed.
Notably,
the {\fontfamily{qcr}\selectfont PERSON} and {\fontfamily{qcr}\selectfont ORG} entities are unique to each instance, rather than across the entire dataset.
Thus,
we prefer random in-instance swapping over in-dataset replacing to avoid introducing irrelevant information from the dataset into the edited evidence.
See examples in Appendix \ref{app:evidence editing}.

\paragraph{Ad-Checking}
The random operation in our editing rules may raise the under-editing evidence (i.e., $\rightarrow$$SUP$) or the over-editing evidence (i.e, $\rightarrow$$NEI$) for $SUP$ instances,
resulting in the generated claim $c'$ based on this evidence being an incorrect semantic perturbation compared to its original claim $c$.
To this end,
we use an existing fact verification model to verify the original claim $c$ based on the edited evidence,
thus ensuring that this evidence is still valid for further providing to the claim \textit{Generator}.
We adopt the RoBERTa \cite{liu2020roberta} model, with the concatenation of the edited evidence and the original claim $c$ as input,
which is fine-tuned on H\textsc{o}V\textsc{er} \cite{hover} dataset with instances labeled as $SUP$, $REF$, and $NEI$.
The edited evidence that yields a $REF$ prediction is retained as counterfactual evidence $E'$ (i.e., $(c,E')$$\rightarrow$$REF$).
If not,
we discard this case for generating counterfactuals.
See Appendix \ref{app:ad-hoc checking} for details.

After Editing and Ad-Checking,
we are ready to proceed with the claim generation for the $SUP$ and $REF$ instances.
We retain the original $REF$ instance as $(c,E,T,R_s,REF)$, and have perturbed the $SUP$ instance $(c,E,T,R_s,SUP)$ to $(c,E',T',R_s,REF)$,
where $T$ and $T'$ denote the set of original and edited causal entities extracted from the token rationales $R_t$, respectively.
Up to this step,
we generate the counterfactuals $(c,E',REF)$ by altering the causal entities within the multi-hop evidence for $(c,E,SUP)$.

\subsection{Generator: Claim Generation}
\label{sec:claim generation}
As \citet{tan-etal-2023-multi2claim} notes,
the direct generation of refuted claims is challenging and may require
additional ontology-like mechanisms to ensure that the generation is plausible but reversed.
Thus,
we opt to generate counterfactual claims $c'$ that are \textbf{supported} by the evidence $E/E'$ from the instances.
Notably, 
we do not intervene too much in its generation process, apart from regulating the generated claim $c'$ sensitive to the causal entities $T/T'$.
This allows us to ensure the linguistically diverse generation while preserving the factual consistency with evidence $E/E'$.

\paragraph{Generation}
We first use a pre-trained generation model (e.g., T5~\cite{raffel2020exploring}) fine-tuned on the $SUP$ instances in FEVEROUS dataset~\cite{feverous},
using the concatenation of all the gold-standard evidence $E$ as input and the corresponding claim $c$ as the target text (i.e., $E \rightarrow c$).
Unlike prior work on editing the original claim $c$, this encourages the linguistically diverse generation by synthesizing the semantic and correlation information between the multi-hop evidence.

Then,
to ensure that the generated claim $c'$ presents factual consistency with the evidence $E/E'$,
we apply constrained beam search decoding \citep{anderson-etal-2017-guided, post-vilar-2018-fast, hu-etal-2019-improved} with entity constraints to guide the claim generation,
by taking the concatenation of all sentence rationales $R_s$ in $E/E'$ as input.

Specifically,
regarding the list of causal entity tokens $dc_i = [t_{i,1}, t_{i,2}, \dots, t_{i,j}]$ within each piece of evidence as disjunctive constraints, 
where $t_{i,*} \in T/T'$ denotes the causal entity in the $i$-th evidence, and 
$j$ is the number of the entities, 
we acquire the conjunction constraint of the beam search by combining of all disjunctive constraints,
\begin{equation}
     CONS = dc_1 \land dc_2 \land \dots \land dc_n,
\end{equation}
where $n$ is the number of evidence. 
The conjunctive constraint during decoding encourages the generated claim $c$ to contain at least one causal entity from each piece of evidence,
thus ensuring factual consistency with the multi-hop evidence.
After repeated generation, we generate $k$ ($k=10$ in our experiments) candidate counterfactual claims $C'=\{c'_1,c'_2,...,c'_k\}$ for each instance.

\paragraph{Post-Checking} 
The claim generation model can be noisy,
potentially leading to the non-reversed predictions of a claim $c'$ given $E$.
To ascertain the label flipping between claim $c'$ and $c$, i.e, $(c'_i|_{i=1}^k,E) \rightarrow y'\neq y$,
by taking the concatenation of each candidate counterfactual claim $c'_i$ with its corresponding original evidence $E$ as input,
we use the same three-way fact verification model as in Ad-Checking to filter the candidate counterfactual claims.
We retain those candidate claims in $C'$ that yield a predicted label $y'\neq y$.  

\paragraph{Discussion} 
Claim generation can also be done by very large language models (LLMs) (e.g., ChatGPT \citep{chatgpt}) with in-context learning \citep{brown2020language,wei2022chain}. 
However, since our editing may introduce inconsistencies with common sense,
we empirically find that the edited evidence $E'$ is more likely to conflict with the internal knowledge of LLMs, 
thus leading to the irrelevant content or even failure in generating the claim $c'$.
Thus, we choose the fine-tuned generation models.

\subsection{Filtering}
\label{sec:filtering}
Unlike prior work that relies on a curated set of minimal edits (e.g., \citet{yang-etal-2021-exploring}),
the strategy in our \textit{Generator} maybe over-generate claim $c'$ with over diverse semantic shift compared to $c$.
Thus, following \citet{paranjape-etal-2022-retrieval},
we use post-hoc filtering with two modules on generated claims $C'$ to ensure the minimal semantic \citep{keane2020good} and topic perturbation compared to the original claim $c$.

\paragraph{Semantic Filtering} The MoverScore \citep{zhao-etal-2019-moverscore}, which combines the contextualized representations with the Earth Mover distance \citep{rubner2000earth}, measures the semantic similarity between two sentences.
We thus use this metric to calculate \emph{semantic fidelity score} between each counterfactual claim in $C'$ and its corresponding original claim $c$,
evaluating the semantic change between these two claims.

\paragraph{Entity Filtering} We introduce the \textit{entity fidelity score} by calculating the overlap rate of entities between strings of claim ($c'$, $c$) pair.
This allows us to ensure topic consistency between $c'$ and $c$, filtering out the irrelevant claims from a topic perspective \cite{si-etal-2021-topic}.

One generated claim $c' \in C'$ with the highest sum score over \emph{semantic fidelity score} and \textit{entity fidelity score} is retained for each instance.
Finally,
our RACE produces the counterfactual data for each instance $(c,E,y)$ in the dataset, including $(c',E,y')$ and $(c,E',y')$.

\begin{table*}[pt]
    \centering
    \resizebox{16cm}{!}{
    \renewcommand\arraystretch{1.1}
\begin{tabular}{lcccccccccc}
\toprule[2pt]
\multirow{2}{*}{\textbf{Source of data}} & \multirow{2}{*}{\textbf{$|D_{train}|$}} & \multicolumn{3}{c}{\textbf{In-domain}}              & \multicolumn{4}{c}{\textbf{Out-of-domain}}                                     & \multicolumn{2}{c}{\textbf{Challenge}} \\
\cmidrule(lr){3-5}\cmidrule(lr){6-9}\cmidrule(lr){10-11}
                                        &                                   & \textbf{H\textsc{o}V\textsc{er}} & \textbf{FEVER} & \textbf{FEVEROUS} & \textbf{PolitiHop} & \textbf{SCIFACT} & \textbf{HealthVer} & \textbf{PubHealth} & \textbf{FM2}    & \textbf{V\textsc{itamin}C}    \\\midrule[1pt]
\textbf{None}                     & 18,171 & 82.55          & 76.70          & 69.43          & 48.74          & 62.77          & 54.98          & 53.01          & 61.51          & 67.05          \\\hdashline
\textbf{EDA}                      & 36,342 & 82.55          & 73.60          & 68.22          & \textbf{54.62}          & 62.77          & 53.68          & 45.99          & 60.56          & 59.63          \\
\hdashline
\textbf{CrossAug}                 & 29,174 & 82.28          & 65.92          & 70.06          & \textbf{54.62}          & 57.98          & 49.24          & 39.15          & 56.12          & 61.66          \\
\textbf{P\textsc{olyjuice}}                & 25,190 & 81.10          & 76.43          & 67.94          & 45.38          & 57.98          & 54.65          & 46.28          & 57.14          & 62.29          \\
\hdashline
\textbf{GPT-3}                     & 24,171 & 80.75 & 72.30 & 67.56 & 51.26 & 64.36 & 49.46 & 42.91 & 61.25 & 62.21          \\
\textbf{ChatGPT}                  & 24,171 & 80.13          & 77.77 & 69.04          & 44.54          & 60.64          & 51.84          & 45.00          & 50.74          & \textbf{68.14} \\
\midrule[0.5pt]
\textit{our} \textbf{RACE (BART)}          & 24,398 & 82.78          & 76.07          & 70.63 & 47.06          & 61.17          & 46.97          & 42.81          & 59.17          & 59.54          \\
\textit{our} \textbf{RACE (GPT-2)}          & 23,645 & 82.53          & 77.15          & 66.07          & 45.38          & \textbf{65.43} & 54.87          & 53.52 & \textbf{62.36} & 67.88          \\
\textit{our} \textbf{RACE (T5-large)}            & 26,638 & \textbf{83.18} & \textbf{78.11}         & \textbf{71.55}         & 47.06          & 62.77 & \textbf{55.84} & \textbf{56.59} & 61.16         & 67.71          \\
\textit{our} \textbf{RACE (T5-base)}            & 26,917 & 83.15 & 75.05          & 70.50          & 52.94          & \textbf{65.43} & 55.41 & 53.52 & 62.19          & 66.50          \\				
\hdashline
\emph{-CONS}                    & 28,468 & 82.53          & 73.93          & 70.09          & 48.74          & 59.04          & 52.71          & 49.06          & 62.28          & 67.31          \\
\emph{-EDIT}                    & 28,359 & 80.75          & 71.50          & 68.13          & 54.62          & 60.64          & 52.92          & 47.47          & 61.33          & 62.72          \\
\emph{-EDIT\&CONS}               & 27,682 & 83.00          & 76.84          & 70.69          & 43.70          & 60.11          & 52.60          & 53.42          & 60.05          & 64.74          \\
\midrule[0.5pt]\midrule[0.5pt]
\textit{w} $(c,E',REF)$                       &                                   &                 &                 &                   &                    &                  &                   &                    &                   &                    \\
\textit{our} \textbf{RACE (BART)}          & 27,909 & \boxed{83.33} & 76.65          & 69.16          & 41.18          & 59.57          & 54.00          & 44.20          & 61.42          & 66.16          \\
\textit{our} \textbf{RACE (GPT-2)}          & 27,156 & 82.78          & 75.31          & 70.52          & 51.26          & 62.77          & 51.62          & 51.83          & 59.97          & 62.18          \\
\textit{our} \textbf{RACE (T5-large)}            & 30,149 & 82.90          & \boxed{78.69}          & 69.29 & 47.90 & 64.89          & 55.41          & 52.03          & 61.08         & 66.31    \\   
\textit{our} \textbf{RACE (T5-base)}            & 30,428 & 82.63          & 76.73          & 70.90 & \boxed{57.14} & 60.11          & 55.63          & 47.87          & 61.33          & 67.66    \\     
\bottomrule[2pt]           
\end{tabular}}
    \caption{
    Fact verification accuracy of various data augmentation methods on different development sets in three settings. $|D_{train}|$ shows the total number of training instances, including 18,171 original H\textsc{o}V\textsc{er} training instances. \textit{w} $(c,E',REF)$ denotes the incorporation of counterfactual instances $(c,E',REF)$ into the training set. \emph{-CONS} denotes the use of beam search instead of constrained beam search in claim generation. \emph{-EDIT} denotes that the evidence editing stage is skipped, and counterfactual claims are generated directly from the original evidence for each original instance. The best of the main results are marked in bold. The results with further improvement in model performance after the incorporation of $(c,E',REF)$ are boxed.}
    \label{tab:generalization}
\end{table*}

\section{Experiments}
\label{sec:experiments}
\paragraph{Datasets}
We generate counterfactual data for H\textsc{o}V\textsc{er}\footnote{Since the H\textsc{o}V\textsc{er} dataset contains explicit multi-hop correlation among evidence based on different reasoning type, 
we choose it to generate counterfactuals and report results in this paper.} training set \citep{hover}, 
a multi-hop dataset with facts sourced from Wikipedia.
We evaluate the model generalization on three types of development sets, (\uppercase\expandafter{\romannumeral1}) In-domain setting (sourced from Wikipedia), including FEVER \citep{fever} and FEVEROUS \citep{feverous}.
(\uppercase\expandafter{\romannumeral2}) Out-of-domain setting (sourced from specific domains), including PolitiHop (political news) \citep{politihop}, SCIFACT (scientific articles) \citep{scifact}, HealthVer \citep{healthver} and PubHealth (public health) \citep{pubhealth}.
(\uppercase\expandafter{\romannumeral3}) Challenge setting (contrastive data), including FM2 \citep{eisenschlos-etal-2021-fool} and V\textsc{itamin}C \citep{schuster-etal-2021-get}. Details and statistics of datasets are presented in Appendix \ref{app:datasets}.

\paragraph{Baselines}
We use three types of baselines to augment the H\textsc{o}V\textsc{er} training set,
(\uppercase\expandafter{\romannumeral1}) Data augmentation method: EDA \citep{wei-zou-2019-eda}. 
(\uppercase\expandafter{\romannumeral2}) Counterfactual data augmentation methods: CrossAug \citep{lee2021crossaug} 
and P\textsc{olyjuice} \citep{wu-etal-2021-polyjuice}. 
(\uppercase\expandafter{\romannumeral3}) LLMs: GPT-3 (\texttt{text-davinci-003}) \citep{brown2020language}
and ChatGPT (\texttt{gpt-3.5-turbo-0301}) \citep{chatgpt}.
More details are presented in Appendix \ref{app:baselines}.

\paragraph{Implementation Details}
In the experiments,
we fine-tune a basic multi-hop fact verification model, an additional RoBERTa-base \citep{liu2020roberta}, on the original training data $(c, E, y)$ and the counterfactual data generated by each method.
The model is evaluated on the development set of different datasets.

For the basic multi-hop fact verification model, we concatenate the claim and all evidence as input sequence, and limit its maximum length to 130. We set the batch size to 4 and optimize the model through a cross entropy loss using the AdamW optimizer \cite{loshchilov2018decoupled} with the learning rate of 1e-5. 
For claim generation, we conduct experiments with four generation models: BART-base \citep{lewis-etal-2020-bart}, T5-base, T5-large \citep{raffel2020exploring} and GPT-2 \citep{radford2019language}.
The beam size is 30 and the max length of generated text is 96. 

\begin{table*}[t]
\centering
\resizebox{\linewidth}{!}{
\renewcommand\arraystretch{1.0}
\begin{tabular}{lp{21cm}}
\toprule[2pt]
\multicolumn{2}{c}{\textbf{Original Instance}} \\\midrule[0.5pt]
\textbf{Claim}& The 1994 British romantic comedy that Charlotte Ninon Coleman played Scarlett in featured the song ``Love''. \\\hdashline
\multirow{4.5}{*}{\textbf{Evidence}} & 1. [Reg Presley] He wrote the song ``Love Is All Around'', which was featured in the films ``Four Weddings and a Funeral'' and ``Love Actually''.  \\ 
& 2. [Charlotte Coleman] Charlotte Ninon Coleman (3 April 1968 – 14 November 2001) was an English actress best known for playing Scarlett in the film ``Four Weddings and a Funera'', Jess in the television drama ``Oranges Are Not the Only Fruit'', and her childhood roles of Sue in ``Worzel Gummidge'' and the character Marmalade Atkins. \\
& 3. [Four Weddings and a Funeral] Four Weddings and a Funeral is a 1994 British romantic comedy film directed by Mike Newell. \\\hdashline
\textbf{Label} & SUPPORTS   \\
\midrule[1pt]
\multicolumn{2}{c}{\textbf{Counterfactual Claims}}  \\\midrule[0.5pt]
\textbf{CrossAug}  & The 1994 British romantic comedy that Charlotte Ninon Coleman played Scarlett in featured the song \color{blue}\st{``Love''}\color{black}.  \\
\textbf{P\textsc{olyjuice}}     & The 1994 British romantic comedy that \textcolor{blue}{did not win} Charlotte Ninon Coleman played Scarlett in featured the song ``Love''.  \\
\textbf{ChatGPT}    & The 1994 \textcolor{blue}{American} romantic comedy that Charlotte Ninon Coleman played Scarlett in featured the song ``Love''. \\
\textit{our} \textbf{RACE (T5-base)} & \textcolor{blue}{Marmalade Atkins directed} the 1948 British romantic comedy that \textcolor{blue}{Reg Presley} played \textcolor{blue}{Charlotte Coleman} in. \textcolor{blue}{It} featured the song ``Love''.  \\                   
\bottomrule[2pt]
\end{tabular}}
    \caption{Examples of counterfactual claims on H\textsc{o}V\textsc{er} training set derived by different methods. The difference between the counterfactual claim and the original claim is highlighted in \textcolor{blue}{blue}. See Table \ref{tab:edited evidence example} in Appendix \ref{app:evidence editing} for the corresponding edited evidence and more examples.
    }
    \label{tab:example}
\end{table*}

\begin{table}[t]
\centering
\resizebox{7.5cm}{!}{
\renewcommand\arraystretch{1.0}
\begin{tabular}{lccccc}
\toprule[2pt]
\textbf{Method}        & \textbf{Flip.} $\uparrow$  & \textbf{Flu.} $\downarrow$  & \textbf{Sim.} $\uparrow$  & \textbf{Div.} $\uparrow$  & \textbf{M.h.} $\uparrow$  \\\midrule[1pt]
\textbf{CrossAug}      & 0.3138  & 209.34 & \textbf{0.6100} & 2.24 & \textbf{0.6090} \\
\textbf{P\textsc{olyjuice}}   & 0.6066   & 195.18  &  0.5969  & 1.20 &  0.5960 \\\hdashline
\textbf{GPT-3}          & 0.3970   &96.84     & 0.5873    & 1.47  & 0.5865 \\
\textbf{ChatGPT}       & 0.4160      & 107.94   & 0.5906 & 1.73  &  0.5898  \\\midrule[0.5pt]
\textbf{RACE (T5-large)} & 0.9402 & \textbf{55.03} & 0.5770  & 11.22 & 0.5763\\
\textbf{RACE (T5-base)} & \textbf{0.9457} & 55.81 & 0.5770  & 11.19 & 0.5763\\
\emph{-FILTER}      & 0.8388   & 58.83   & 0.5773  & \textbf{13.01}   &0.5766  \\
\bottomrule[2pt]
\end{tabular}}
    \caption{
    Automatic intrinsic evaluation results. For \textbf{Flip rate (Flip.)}, we use a RoBERTa-based classifier fine-tuned on the H\textsc{o}V\textsc{er} training set to calculate the verification accuracy of the instance $(c',E,y')$. For \textbf{Fluency (Flu.)}, following previous work \citep{atanasova-etal-2020-generating, he2023reinforcement}, we use the perplexity scores calculated by GPT-2 to evaluate the fluency of $c'$. For \textbf{Similarity (Sim.)}, we calculate the MoverScore between $c'$ and $c$. For \textbf{Diversity (Div.)}, following \citet{rani2023factify}, we use the inverse of the BLEU score \citep{papineni-etal-2002-bleu} to measure dissimilarity between $c'$ and $c$. For \textbf{Multi hop (M.h.)}, we employ MoverScore to calculate the average semantic similarity between $c'$ and $e_i$, where $e_i \in E$, to evaluate the coherence like \citet{he2023reinforcement}. \emph{-FILTER} denotes the evaluation of all the generated claims before post-checking and filtering stage. The best results are marked in bold.}
    \label{tab:intrinsic}
\end{table}

\section{Results and Discussion}
\label{sec:results}

\subsection{Main Results}
\label{subsec:main_results}
Neglecting the logical relationships within the correlated input results in a failure to generate counterfactual evidence $E'$ for baselines. Thus, we mainly compare the effects of the counterfactual data $(c', E, y')$ generated by the different methods. Meanwhile, we also report the results after incorporating  $(c,E',REF)$ into the training set (bottom of Table \ref{tab:generalization}).

\paragraph{Out-of-domain Setting}
Table \ref{tab:generalization} shows the effects of the data generated by RACE and baselines on the OOD generalization.
We can observe that,
(\uppercase\expandafter{\romannumeral1}) RACE significantly improves model performance on PolitiHop, SCIFACT and PubHealth compared to the results without data augmentation,
and outperforms baselines on almost all OOD datasets, demonstrating the effectiveness of our augmentation strategy for multi-hop fact verification task. 
(\uppercase\expandafter{\romannumeral2}) 
RACE significantly outperforms P\textsc{olyjuice}, showing that the general-purpose CDA method, designed for tasks without requiring complex reasoning on the input, fails to achieve acceptable results on multi-hop fact verification task, and even impairs the OOD generalization. 
(\uppercase\expandafter{\romannumeral3}) The counterfactual data generated by LLMs provides little improvement in OOD generalization, demonstrating that CDA for multi-hop fact verification task remains challenging for LLMs by using the in-context learning alone.
(\uppercase\expandafter{\romannumeral4}) The incorporation of $(c,E',REF)$ further improves the model generalization to a certain extent on PolitiHop, indicating that the edited evidence still remains multi-hop correlated and reasonable.

\paragraph{Challenge Setting}
Comparing the results on challenging contrastive datasets, as Table \ref{tab:generalization} shows, training with RACE data improves the fact verification accuracy,
while almost all the baselines degrade the performance of the model.
This phenomenon confirms that our method improves model robustness to spurious correlations.
Additionally,
the incorporation of the $(c,E',REF)$ yields no improvement in verification accuracy, probably because these datasets are constructed in response to the elimination of spurious correlations between features in \textbf{claim} and labels.

\paragraph{In-domain Setting}
As shown in Table \ref{tab:generalization}, RACE improves the model performance on in-domain data, while most baselines tend to degrade it. Notably, our method has the most significant improvement on the FEVEROUS development set, which requires four pieces of true evidence to verify each claim on average. This further demonstrates the effectiveness of our method for multi-hop fact verification task.

\subsection{Ablation Study}
We conduct ablation studies on evidence editing and claim generation stage to verify the reasonableness of causal entities in token rationales. 
All the experiments are conducted based on RACE (T5-base).

Firstly, we use ordinary beam search instead of constrained beam search during the claim generation stage (i.e., \emph{-CONS} in Table \ref{tab:generalization}). The results in Table \ref{tab:generalization} reveal that a significant performance decrease occurs on both in-domain and OOD data.
It might be explained by constraints based on entities in token rationales, which allow the generated claim to be multi-hop and topic consistent with the original claim, resulting in a more efficient counterfactual. 
In contrast, we note a slight improvement on the challenge datasets, which might be attributed to the shorter length of claims in both datasets (each claim contains about 13 words on average).

Then, we skip the evidence editing stage and directly generate the counterfactual claims for all the instances (i.e, \emph{-EDIT} in Table \ref{tab:generalization}) by a T5-base language model.
The model is fine-tuned on FEVEROUS to generate claims that are supported or refuted by the input evidence via setting the prefix.
As shown in Table \ref{tab:generalization}, the accuracy decreases substantially on almost all datasets, except for PolitiHop. It can be explained by the fact that political news typically focuses on event information rather than entity information, hence entity-based evidence editing fails to improve model generalization on PolitiHop.

Finally, we further remove both the constrained beam search and evidence editing stage (i.e., \emph{-CONS\&EDIT} in Table \ref{tab:generalization}). A significant decrease in accuracy is observed on both OOD and challenge data, which demonstrates that the proposed evidence editing based on rationales and claim generation based on entities are crucial for improving the generalization and robustness of the multi-hop fact verification models.

\subsection{Intrinsic Evaluation}
\label{subsec:intrinsic_evaluation}

For further analysis of the quality of the generated counterfactual claims, following \citet{chemmengath-etal-2022-cat} and \citet{dixit-etal-2022-core}, we automatically and manually evaluate the generated counterfactual claims according to the following five criteria: 
(\uppercase\expandafter{\romannumeral1}) \emph{Flip rate (Flip.)}, measuring if the label of the generated claim is flipped based on the original evidence;
(\uppercase\expandafter{\romannumeral2}) \emph{Fluency (Flu.)}, measuring whether the generated claim is grammatically correct and semantically meaningful; 
(\uppercase\expandafter{\romannumeral3}) \emph{Diversity (Div.)}, reflecting the linguistic diversity of the generated claim compared to the original claim; 
(\uppercase\expandafter{\romannumeral4}) \emph{Similarity (Sim.)}, measuring the degree of semantic similarity between the generated claim and the original claim, where we use MoverScore \citep{zhao-etal-2019-moverscore} instead of Levenshtein edit distance \citep{levenshtein1966binary} in the automatic evaluation to balance with diversity;
(\uppercase\expandafter{\romannumeral5}) \emph{Multi hop (M.h.)}, measuring whether the generated claim is multi-hop and relevant to the evidence.

\paragraph{Automatic Evaluation}
For a fair comparison, the claims generated before and after the post-checking and filtering are compared with the baselines separately. As shown in Table \ref{tab:intrinsic}, 
RACE outperforms baselines significantly in terms of flip rate, diversity, and fluency. It demonstrates the ability of RACE to generate fluent and \emph{linguistically diverse} counterfactual claims based on the edited evidence,
while keeping \emph{label flipping} and \emph{logical relationships} with the original evidence.
Moreover, the counterfactual claim after the filtering stage achieves a higher flip rate and fluency score compared to the one before filtering, which illustrates the necessity of the filtering stage for generating high-quality counterfactual data.  
For automatic evaluation of \textit{Multi hop}, we follow \citet{he2023reinforcement} to use MoverScore to evaluate the multi hop of counterfactuals. And all methods achieve comparable results. However, we argue that this is compromised since its solely semantic comparison cannot reflect whether all the evidence can be aggregated as a whole to verify the counterfactual claim.

\begin{figure}[t]
    \centering
    \includegraphics[width=8cm]{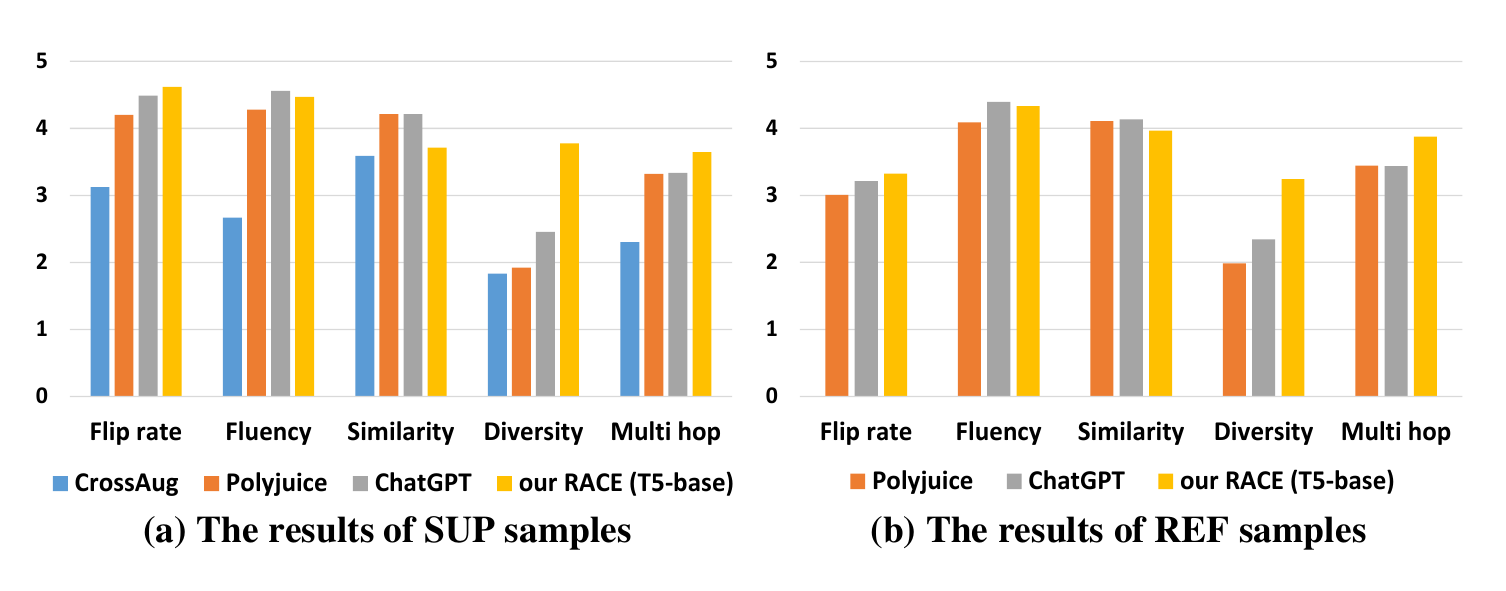}
    \caption{The results of human evaluation, where 1 indicates a complete breach of the criteria and 5 indicates full compliance. The inter-rate agreement measured by Krippendorff's $\alpha$ \citep{krippendorff2011computing} is 0.54.}
    \label{fig:human evaluation}
\end{figure}

\paragraph{Manual Evaluation}
To address the limitations of the automatic evaluation, we adopt the human evaluation to qualify the counterfactuals from different aspects. Specifically, we randomly select 30 $SUP$ instances and 30 $REF$ instances and ask three postgraduate students with an NLP background to score counterfactual claims in a likert scale of 1 to 5 according to the above criteria. 
Since CrossAug can only generate counterfactuals for $SUP$ instances, we compare the results on $SUP$ and $REF$ instances separately. 
The evaluation results are shown in Figure \ref{fig:human evaluation}. It can be observed that RACE well outperforms baselines, particularly in terms of diversity, which illustrates the ability of RACE to generate \emph{human-readable}, \emph{diverse}, and \emph{label-flipping} counterfactual claims.
Meanwhile, entity constraint-based generation enables RACE to generate multi-hop claims.

Overall, both the automatic and manual evaluation results show the effectiveness of RACE from different aspects for multi-hop fact verification task.

\subsection{Qualitative Evaluation}
\label{subsec:qualitative_evaluation}
Table \ref{tab:example} presents an example of the original instance and the counterfactual claims generated by different methods. The words that differ from the original claim are highlighted. 
It can be observed that RACE generates a linguistically diverse and fluent counterfactual claim, and the original label is successfully flipped. Obviously, the counterfactual claim generated by RACE can be combined with the original evidence to form a valid multi-hop fact verification instance, which is logical and can be verified according to the given evidence.
Moreover, the claim generated by RACE is semantically and lexically similar to the original claim, benefiting casual entities in multi-hop rationales.
Nevertheless, the baselines tend to simply modify the original claim, despite the use of language models.  
As shown in Table \ref{tab:example}, most of the baselines (including LLMs), prefer to add ``not'' to the original claim or make antonym substitutions.
Such modifications make the counterfactual claims lexically similar to the original claim, but are not valid for multi-hop fact verification and cannot generate a diverse and logical counterfactual claim (as evidenced by lower flip rate and diversity in Table \ref{tab:intrinsic} and Figure \ref{fig:human evaluation}). 

\subsection{Effect of Generation Models}
\label{subsec:generation_models}
We adopt different generation models to test the effect of the generation ability on our method, which aims to illustrate the independence of our proposed method from a particular generation model (i.e., Generation Model-Agnostic). As shown in Table \ref{tab:generalization}, compared to the baselines, our RACE yields a comparable or improved performance based on different generation models, especially the results based on T5-base and T5-large. Besides, We empirically find that different generation models have more prominent performance on specific datasets, e.g., GPT-2 on SCIFACT and FM2 datasets, and T5 on 6 datasets.

To explore the effect of the number of parameters, we further compare the results based on T5-base and T5-large. As Table \ref{tab:intrinsic} and \ref{tab:generalization} shows, compared to T5-base, counterfactuals generated by fine-tuned T5-large are more fluent and linguistically diverse, and further improve the model performance on most datasets. This illustrates that it is possible to further improve the effectiveness of our method by using a more powerful generation model. Thus, for the choice of the generation model, we recommend choosing the powerful possible generation model in the absence of the priors to the data.

\section{Conclusion}
We present a novel rationale-sensitive pipeline counterfactual data augmentation method (RACE) to generate \textit{logical}, \textit{diverse}, and \textit{label-flipping} counterfactuals for multi-hop fact verification task. 
An Explain-Edit-Generate architecture is constructed to generate diverse and logical counterfactual claims based on the rationales.
Then, a filter process with two modules is employed to further regularize semantic and topic consistency.
Experimental results reveal the
improvement in OOD generalization and robustness of the proposed method. 
Intrinsic evaluation and qualitative evaluation of counterfactual claims show that RACE can generate linguistically diverse and label-flipping counterfactual data while preserving logical relationships.

\section*{Limitations}
As multi-hop fact verification is a relatively complex reasoning task, designing an effective method to generate counterfactuals for this task requires a consideration of the logical relationships between the claim and the evidence and between multiple pieces of evidence, making our proposed method more complex and cumbersome. 
Meanwhile, the use of heuristic rules in the editing process results in the inability to generalize to other tasks and the need to recreate the rules. 
In addition, the prompts given to LLMs for generating counterfactual claims can be further elaborated, e.g., using chain-of-thought, to exploit more potential of LLMs on CDA for multi-hop fact verification task.

In the future, due to the flexible generation of LLMs, we will explore the construction of effective prompts to generate counterfactuals for multi-hop fact verification using the Chain-of-Thought.

\section*{Acknowledgement}

The authors would like to thank the anonymous
reviewers for their insightful comments. This work
is funded by the National Natural Science Foundation of China (62176053) and supported by the Big Data Computing Center of Southeast
University. YH is supported by a Turing AI Fellowship funded by the UK Research and Innovation (grant no. EP/V020579/1, EP/V020579/2).

\bibliography{emnlp2023}
\bibliographystyle{acl_natbib}

\appendix

\begin{table*}[t]
\centering
\resizebox{\linewidth}{!}{
\renewcommand\arraystretch{1.1}
\begin{tabular}{lp{20cm}}
\toprule[2pt]
\multicolumn{2}{c}{\textbf{Original Instance}} \\\midrule[0.5pt]
\textbf{Claim}& The 1994 British romantic comedy that Charlotte Ninon Coleman played Scarlett in featured the song ``Love''. \\\hdashline
\multirow{4.5}{*}{\textbf{Evidence}} & 1. [Reg Presley] He wrote the song ``Love Is All Around'', which was featured in the films ``Four Weddings and a Funeral'' and ``Love Actually''.  \\ 
& 2. [Charlotte Coleman] Charlotte Ninon Coleman (3 April 1968 – 14 November 2001) was an English actress best known for playing Scarlett in the film ``Four Weddings and a Funera'', Jess in the television drama ``Oranges Are Not the Only Fruit'', and her childhood roles of Sue in  ``Worzel Gummidge'' and the character Marmalade Atkins. \\
\renewcommand\arraystretch{1.2}
& 3. [Four Weddings and a Funeral] Four Weddings and a Funeral is a 1994 British romantic comedy film directed by Mike Newell. \\\hdashline
\textbf{Label} & SUPPORTS   \\
\midrule[1pt]
\multicolumn{2}{c}{\textbf{Edited Evidence}}  \\\midrule[0.5pt]
\multirow{6.8}{*}{\textbf{Edired Evidence}} & 1. [\textcolor{blue}{Mike Newell}] He wrote the song ``Love Is All Around'', which was featured in the films ``Four Weddings and a Funeral'' and ``Love Actually''.  \\ 
& 2. [\textcolor{blue}{Reg Presley}] \textcolor{blue}{Reg Presley} \textcolor{blue}{(3 August 1987 – 26 June 2000)} was an English actress best known for playing \textcolor{blue}{Charlotte Coleman} in  the film ``Four Weddings and a Funera'', \textcolor{blue}{Scarlett} in the television drama ``Oranges Are Not the Only Fruit'', and her childhood roles of Sue in ``Worzel Gummidge'' and the character \textcolor{blue}{Jess}. \\
& 3. [Four Weddings and a Funeral] Four Weddings and a Funeral is a \textcolor{blue}{1948} British romantic comedy film directed by \textcolor{blue}{Marmalade Atkins}. \\
\midrule[1pt]
\multicolumn{2}{c}{\textbf{Counterfactual Claims}}  \\\midrule[0.5pt]
\textbf{CrossAug}  & The 1994 British romantic comedy that Charlotte Ninon Coleman played Scarlett in featured the song \color{blue}\st{``Love''}\color{black}.  \\
\textbf{P\textsc{olyjuice}}     & The 1994 British romantic comedy that \textcolor{blue}{did not win} Charlotte Ninon Coleman played Scarlett in featured the song ``Love''.  \\
\textbf{ChatGPT}    & The 1994 \textcolor{blue}{American} romantic comedy that Charlotte Ninon Coleman played Scarlett in featured the song ``Love''. \\
\multirow{2}{*}{\textit{our} \textbf{RACE}} & \textcolor{blue}{Marmalade Atkins directed} the 1948 British romantic comedy that \textcolor{blue}{Reg Presley} played \textcolor{blue}{Charlotte Coleman} in. \textcolor{blue}{It} featured the song ``Love''.  \\                   
\bottomrule[2pt]
\toprule[2pt]
\multicolumn{2}{c}{\textbf{Original Instance}} \\\midrule[0.5pt]
\textbf{Claim}& Bruce Geller who died in 1978 developed American television detective show Mannix. \\\hdashline
\multirow{3.5}{*}{\textbf{Evidence}} & 1. [Mannix] Created by Richard Levinson and William Link and developed by executive producer Bruce Geller, the title character, Joe Mannix, is a private investigator.  \\ 
& 2. [Bruce Geller] Bruce Bernard Geller (October 13, 1930 – May 21, 1978) was an American lyricist, screenwriter, director, and television producer. \\
\hdashline
\textbf{Label} & SUPPORTS   \\
\midrule[1pt]
\multicolumn{2}{c}{\textbf{Edited Evidence}}  \\\midrule[0.5pt]
\multirow{3.6}{*}{\textbf{Edired Evidence}} & 1. [Mannix] Created by \textcolor{blue}{Joe Mannix} and Richard Levinson and developed by executive producer \textcolor{blue}{William Link}, the title character, \textcolor{blue}{Bruce Geller}, is a private investigator.  \\ 
& 2. [\textcolor{blue}{William Link}] \textcolor{blue}{William Link (December 14, 1898 – April 30, 1977)} was an American lyricist, screenwriter, director, and television producer. \\
\midrule[1pt]
\multicolumn{2}{c}{\textbf{Counterfactual Claims}}  \\\midrule[0.5pt]
\textbf{CrossAug}  & Bruce Geller who \textcolor{blue}{passed away} in 1978 developed American television detective show Mannix.  \\
\textbf{P\textsc{olyjuice}}     & Bruce Geller who died in 1978, \textcolor{blue}{did not} developed American television detective show Mannix.  \\
\textbf{ChatGPT}    & Bruce Geller, who \textcolor{blue}{passed away} in \textcolor{blue}{1985}, developed the American television detective show Mannix. \\
\multirow{2}{*}{\textit{our} \textbf{RACE}} & \textcolor{blue}{The executive producer of} American television detective show Mannix died in \textcolor{blue}{1877}. \textcolor{blue}{The show was created by Joe Mannix and Richard Levinson.}  \\                   
\bottomrule[2pt]
\end{tabular}}
    \caption{Examples of edited evidence and counterfactual claims on H\textsc{o}V\textsc{er} training set. The differences from the original instance are highlighted in \textcolor{blue}{blue}. 
    }
    \label{tab:edited evidence example}
\end{table*}

\section{Evidence Editing}
\label{app:evidence editing}
Table \ref{tab:edited evidence example} shows examples of the evidence edited by RACE. 
We can observe that rationale- and entity-based editing enables the edited evidence to still retain multi-hop correlation with each other and present a completely different fact from the original evidence.
Hence, the claim generator can generate logical, fluent, and linguistically diverse counterfactual claims based on the edited evidence.

\begin{table}[t]
\resizebox{\linewidth}{!}{
\renewcommand\arraystretch{1.2}
\begin{tabular}{ccccc}
\toprule[2pt]
\textbf{Augmented H\textsc{o}V\textsc{er}} & \textbf{Num.SUP} & \textbf{Num.REF} & \textbf{Num.NEI} & \textbf{Total} \\\midrule[1pt]
\textbf{Train}      & 11,023        & 7,148        & 9,086          & 27,572          \\
\textbf{Dev}        & 2,000         & 2,000        & 2,000          & 6,000            \\
\bottomrule[2pt]
\end{tabular}}
\caption{The statistics of augmented H\textsc{o}V\textsc{er} with $NEI$ instances. Num.SUP, Num.REF and Num.NEI are the number of $SUP$ instances, $REF$ instances, and $NEI$ instances, respectively.}
\label{tab:augmented hover}
\end{table}

\section{Checking Module}
\label{app:ad-hoc checking}
For the ad- and post-checking module, we fine-tune a RoBERTa-base classifier to filter invalid edited evidence and counterfactual claims, respectively. To improve the quality of the retained data, we fine-tune it on the $SUP$, $REF$, and $NEI$ instances rather than just the $SUP$ and $REF$ instances. 

Considering that we perform CDA on H\textsc{o}V\textsc{er} training set during the experiment while no $NEI$ instances are available in H\textsc{o}V\textsc{er}, 
we first conduct data augmentation on H\textsc{o}V\textsc{er} dataset to incorporate $NEI$ instances by perturbing existing instances. 
Specifically, for a random instance in H\textsc{o}V\textsc{er}, we randomly remove one piece of true evidence or randomly pair the claim with the evidence of another instance. 
To avoid imbalance classes, we randomly select half of the $SUP$ instances and half of the $REF$ instances for perturbation and each perturbation strategy is employed with equal probability.
Finally, the fine-tuned RoBERTa-base classifier has 81.23\% on label accuracy of claim verification on $NEI$ augmented H\textsc{o}V\textsc{er} development set. The statistics of $NEI$ augmented H\textsc{o}V\textsc{er} are shown in Table \ref{tab:augmented hover}.

Other implementation details are the same as the fact verification model in the OOD generalization experiment described in Section \ref{sec:experiments}. 

\begin{table*}[t]
\centering
\resizebox{\linewidth}{!}{
\renewcommand\arraystretch{1.2}
\begin{tabular}{p{20cm}}
\toprule[2pt]
Given an original claim with corresponding evidence and label (SUPPORTS or REFUTES), generate a counterfactual claim based on the evidence, taking care to ensure that the generated counterfactual claim is as \textbf{similar} as possible to the original claim, while being aware of linguistic \textbf{diversity} and the \textbf{change} of labels. \\
\textbf{Example:}    \\
\textbf{Claim:} Bettany Hughes, an English historian scholar, born May 15th, 1967, presented "The Spartans".       \\
\textbf{Evidence:}     \\
The Spartans (documentary): `` The Spartans '' was a 3-part historical documentary series first broadcast on UK terrestrial Channel 4 in 2003, presented by Bettany Hughes.      \\
Bettany Hughes: Bettany Hughes ( born May 15 , 1967 ) is an English historian, author, and broadcaster.         \\
\textbf{Label:} SUPPORTS          \\
\textbf{Generate a counterfactual claim: }\\
``The Spartans'' is a documentary presented by Bettany Hughes, an American historian scholar born on March 24, 1980.  \\
\\
\textbf{Claim:} The writer Norman Alfred William Lindsay enjoyed boxing, but the author of The Hundred Secret Senses did not.\\
\textbf{Evidence:}   \\
Amy Tan: Amy Tan ( born February 19, 1952 ) is an American writer whose works explore mother-daughter relationships and the Chinese American experience.  \\
The Hundred Secret Senses: The Hundred Secret Senses is a bestselling 1995 novel by Chinese-American writer Amy Tan.  \\
Norman Lindsay: Norman Alfred William Lindsay ( 22 February 1879 – 21 November 1969 ) was an Australian artist, etcher, sculptor, writer, editorial cartoonist, scale modeller, and an accomplished amateur boxer.  \\
\textbf{Label:} SUPPORTS \\
\textbf{Generate a counterfactual claim: }                     \\
\bottomrule[2pt]
\end{tabular}}
\caption{An example of prompt given to GPT-3 and ChatGPT for generating counterfactual claims.}
\label{tab:prompt}
\end{table*}

\begin{table}[t]
\centering
\resizebox{\linewidth}{!}{
\renewcommand\arraystretch{1.2}
\begin{tabular}{lccc}
\toprule[2pt]
\textbf{Dataset}       & \textbf{Num.SUP} & \textbf{Num.REF} & \textbf{Total} \\
\midrule[1pt]
\textbf{H\textsc{o}V\textsc{er} Dev}     & 2,000             & 2,000             & 4,000           \\
\textbf{FEVER Dev}     & 6,666             & 6,666             & 13,332          \\
\textbf{FEVEROUS Dev}  & 3,908             & 3,481             & 7,389           \\
\textbf{PolitiHop Dev} & 21               & 98               & 119            \\
\textbf{SCIFACT Dev}   & 124              & 64               & 188            \\
\textbf{HealthVer Dev} & 533              & 391              & 924            \\
\textbf{PubHealth Dev} & 628              & 544              & 1,172           \\
\textbf{FM2 Dev}       & 596              & 573              & 1,169           \\
\textbf{V\textsc{itamin}C Dev}      & 31,484            & 22,528            & 54,012  \\
\bottomrule[2pt]
\end{tabular}}
\caption{The statistics of the datasets we used in our experiments.}
\label{tab:statistics}
\end{table}

\section{Datasets}
\label{app:datasets}
\begin{itemize}
    \item \textbf{H\textsc{o}V\textsc{er}} \citep{hover}, a dataset for multi-hop fact verification, which challenges models to extract relevant evidence from several Wikipedia articles and verify whether the claim is SUPPORTED or REFUTED by the evidence. We construct the dataset following \citet{baleen}, where each claim is associated with five pieces of evidence.
    \item \textbf{FEVER} \citep{fever}, a large-scale fact verification dataset with the claims generated by altering sentences extracted from Wikipedia. The claims in FEVER are classified as SUPPORTS, REFUTES or NOT ENOUGH INFO (NEI) by annotators and more than 87\% of them only require information from a single Wikipedia article. We remove the instances with NEI label and only retain the other two classes of instances in our experiments.
    \item \textbf{FEVEROUS} \citep{feverous}, a large-scale multi-hop fact verification dataset consisting of claims verified against Wikipedia pages and labeled as SUPPORTS, REFUTES or NOT ENOUGH INFO (NEI). Each claim has evidence in the form of sentences and/or cells from tables on Wikipedia. Following \citet{Chen2020TabFact:} and \citet{si2022exploring}, we employ the simple table linearization template to generate contextualized sequence representations for table evidence. We remove the instances with NEI label and only retain the other two classes of instances in our experiments.
    \item \textbf{PolitiHop} \citep{politihop}, a multi-hop fact verification dataset of real-world claims with manual annotations of evidence from PolitiFact articles. The labels include FALSE, HALF-TRUE and TRUE. In our experiments, we remove the instances with HALF-TRUE label and only retain the other two classes of instances.
    \item \textbf{SCIFACT} \citep{scifact}, a scientific fact verification dataset of 1.4K expert-written scientific claims paired with evidence. As with the above dataset, we only retain the instances with SUPPORTS and REFUTES labels to evaluate the model.
    \item \textbf{HealthVer} \citep{healthver}, an evidence-based fact verification dataset for health-related claims, where the relations between each piece of evidence and the associated claim are manually annotated as SUPPORT, REFUTE, and NEUTRAL. We remove the instances with the NEUTRAL label. As the evidence provided by HealthVer contains several sentences, we split it into multiple pieces of evidence to simulate a multi-hop scenario.
    \item \textbf{PubHealth} \citep{pubhealth}, a 4-way classification dataset for explainable fact verification with gold standard explanations by journalists in the public health setting. We only retain the instances with TRUE and FALSE labels, and the explanation provided is split into separate sentences as multiple pieces of evidence.
    \item \textbf{FM2} \citep{eisenschlos-etal-2021-fool}, a large-scale dataset of challenging claim-evidence pairs collected through a fun multi-player game which encourages adversarial instances and drastically lowers the number of the instances with ``shortcuts''. All the claims need to be verified $\in$ \{SUPPORTS, REFUTES\}.
    \item \textbf{V\textsc{itamin}C} \citep{schuster-etal-2021-get}, a large-scale contrastive fact verification dataset, where each contrastive claim is manually written by annotators based on Wikipedia revisions. We only retain the instances with SUPPORTS and REFUTES labels in our experiments.
\end{itemize}
We only test the performance of the basic multi-hop fact verification model on the development set of the above datasets in our experiments. The statistics are shown in Table \ref{tab:statistics}.

\section{Baselines}
\label{app:baselines}
In our experiments, we compare our method with the following baselines.
\begin{itemize}
    \item \textbf{EDA} \citep{wei-zou-2019-eda}, a data augmentation method that applies four simple operations, including synonym replacement, random insertion, random swap, and random deletion, to original sentences to generate new instances.
    \item \textbf{CrossAug} \citep{lee2021crossaug}, a counterfactual data augmentation method that employs a two-stage augmentation pipeline to generate contrastive claims and evidence from existing $SUP$ instances.
    \item \textbf{P\textsc{olyjuice}} \citep{wu-etal-2021-polyjuice}), a general-purpose counterfactual generator based on fine-tuned GPT-2 that allows for control over perturbation types and locations.
    \item \textbf{GPT-3} (\texttt{text-davinci-003}) \citep{brown2020language}, a large autoregressive language model with superb few-shot and in-context learning capabilities.
    \item \textbf{ChatGPT} (\texttt{gpt-3.5-turbo-0301}) \citep{chatgpt}, a powerful GPT-3 based model which is trained to follow an instruction in a prompt and provide a detailed response.
\end{itemize}

For EDA\footnote{\url{https://github.com/jasonwei20/eda_nlp}} and CrossAug\footnote{\url{https://github.com/minwhoo/CrossAug}}, all the experimental setups of them are followed from the original papers and 
all hyperparameters are set to the same values as in the official code.
For P\textsc{olyjuice}\footnote{\url{https://github.com/tongshuangwu/polyjuice}}, we set the control code to ``negation'', the beam size to 10, and generate one counterfactual claim for each original claim.
All the inputs to the above baselines are only the original claim.

For GPT-3 and ChatGPT, we make use of the APIs provided by OpenAI\footnote{\url{https://openai.com/product}} for generating counterfactual claims and design a prompt with a task introduction and demonstration as input, as shown in the Table \ref{tab:prompt}.

\end{document}